\documentclass[runningheads]{llncs}
\usepackage[T1]{fontenc}
\usepackage{graphicx}
\usepackage{booktabs}
\usepackage[misc]{ifsym}
\newcommand{\corr}{(\Letter)}
\usepackage{mwe}
\usepackage{multirow}%
\usepackage{amsmath,amssymb,amsfonts}%
\usepackage{mathrsfs}%

\usepackage{xcolor}%
\usepackage{textcomp}%
\usepackage{manyfoot}%
\usepackage[noline,ruled,vlined]{algorithm2e}
\usepackage{listings}
\usepackage{hyperref}
\usepackage{cleveref}

\providecommand{\inner}[2]{#1\trans#2} %{\langle #1, #2 \rangle} %{#1\trans #2}
\providecommand{\with}{\!\mid\,}
\providecommand{\zeronorm}[1]{\|#1\|_0}
\providecommand{\onenorm}[1]{\|#1\|_1}
\providecommand{\twonorm}[1]{\|#1\|_2}
\providecommand{\trans}{^\mathrm{T}}
\providecommand{\step}{\mathbf{1}}
\providecommand{\bq}{\mathbf{q}}
\providecommand{\bw}{\mathbf{w}}
\providecommand{\bteta}{\boldsymbol{\theta}}
\providecommand{\bx}{\mathbf{x}}
\providecommand{\bt}{\mathbf{t}}
\providecommand{\bg}{\mathbf{g}}
\providecommand{\by}{\mathbf{y}}
\providecommand{\bX}{\mathbf{X}}
\providecommand{\bbeta}{\boldsymbol{\beta}}
\providecommand{\bTeta}{\boldsymbol{\Theta}}
\providecommand{\bsigma}{\boldsymbol{\sigma}}
\providecommand{\bbR}{\mathbb{R}}
\providecommand{\setQ}{\mathcal{Q}}
\providecommand{\setW}{\mathcal{W}}
\providecommand{\setG}{\mathcal{G}}
\providecommand{\setI}{\mathcal{I}}
\providecommand{\loss}{\ell}
\providecommand{\Risk}{\mathcal{R}}
\providecommand{\obj}{\mathrm{obj}}
\providecommand{\bobj}{\mathbf{obj}}
\providecommand{\gb}{\mathrm{gb}}
\providecommand{\gs}{\mathrm{gs}}
\providecommand{\log}{\mathrm{log}}
\providecommand{\sqr}{\mathrm{sqr}}
\providecommand{\E}{\mathrm{E}}
\providecommand{\np}{k_{\mathrm{pr}}}
\providecommand{\ns}{k_{\mathrm{sp}}}
\providecommand{\halfspace}{\hspace{0.1em}}
\providecommand{\thickhline}{\noalign{\hrule height 0.25mm}}

\providecommand{\extratiny}[1]{\scalebox{0.8}{#1}}
\providecommand{\extratinytiny}[1]{\scalebox{0.72}{#1}}

\providecommand{\given}{\,|\,}
\providecommand{\meduplo}[3]{\extratinytiny{#3} & #1 & \extratinytiny{#2} }

\providecommand{\sign}{\mathrm{sign}}
\providecommand{\signg}{\zeta}
\DeclareMathOperator*{\argmax}{\mathrm{arg\,max}}
\DeclareMathOperator*{\argmin}{\mathrm{arg\,min}}
\usepackage{adjustbox} 
\usepackage{caption}
\usepackage{subcaption}
\usepackage{array}
\usepackage{booktabs}
\usepackage{rotating}
\usepackage{array}
\usepackage{tabularx}

\begin{document}

\title{Interpretable Representation Learning for Additive Rule Ensembles}

\titlerunning{Interpretable Representation Learning for Additive Rule Ensembles}
% If the full title of your paper is short enough to also fit in the running head, you can omit the abbreviated paper title here. You can check as follows: if you comment out the \titlerunning line, something will appear in the header of all odd-numbered pages of your PDF from page 3 onward. This something is either the full title (in which case all is well), or the error message "Title Suppressed Due to Excessive Length". If this error message appears, you're going to want to provide an abbreviated title within the \titlerunning command, because if you won't do it, Springer will do it for you.

%N.B.: Author information (both in the \author{} and \authorrunning{} command) should only be present in the Camera-Ready Version of your paper. The version that you initially submit for review, ought to be double-blind. So, when initially submitting your paper, use:
% \author{Author information scrubbed for double-blind reviewing}

\author{Shahrzad Behzadimanesh\inst{1} \corr \and
Pierre Le Bodic\inst{1} \and
Geoffrey I. Webb\inst{1} \and 
Mario Boley \inst{1,2}}

% You may leave out the orcidID information, if you want to.
% Use \corr to indicate the corresponding author. Note the spacing around the \corr command. Only one author can be the corresponding author.

%N.B.: comment out the \authorrunning{} command for the double-blind version of your paper submitted for review. Later, if your paper is accepted, use the command for the Camera-Ready Version.
\authorrunning{S. Behzadimanesh et al.}
% If there are two authors, write 'A.L. Benjamin and C.C. Broadus Jr.'
% If there are more than two authors, '[...] et al.' is used.

\institute{Department of Data Science and Artificial Intelligence, Monash University \email{\{shahrzad.behzadimanesh,pierre.lebodic,geoff.webb\}@monash.edu}
\and
Department of Information Systems, University of Haifa, \email{mboley@is.haifa.ac.il}}

\maketitle              % typeset the header of the contribution

\begin{abstract}

Small additive ensembles of symbolic rules offer interpretable prediction models. Traditionally, these ensembles use rule conditions based on conjunctions of simple threshold propositions $x \geq t$ on a single input variable $x$ and threshold $t$, resulting geometrically in axis-parallel polytopes as decision regions. While this form ensures a high degree of interpretability for individual rules and can be learned efficiently using the gradient boosting approach, it relies on having access to a curated set of expressive and ideally independent input features so that a small ensemble of axis-parallel regions can describe the target variable well. Absent such features, reaching sufficient accuracy requires increasing the number and complexity of individual rules, which diminishes the interpretability of the model.
Here, we extend classical rule ensembles by introducing logical propositions with learnable sparse linear transformations of input variables, i.e., propositions of the form $\bx\trans\bw \geq t$, where $\bw$ is a learnable sparse weight vector, enabling decision regions as general polytopes with oblique faces. We propose a learning method using sequential greedy optimization based on an iteratively reweighted formulation of logistic regression. 
Experimental results demonstrate that the proposed method efficiently constructs rule ensembles with the same test risk as state-of-the-art methods while significantly reducing model complexity across ten benchmark datasets.


\end{abstract}

\providecommand{\inner}[2]{#1\trans#2} %{\langle #1, #2 \rangle} %{#1\trans #2}
\providecommand{\with}{\!\mid\,}
\providecommand{\zeronorm}[1]{\|#1\|_0}
\providecommand{\onenorm}[1]{\|#1\|_1}
\providecommand{\twonorm}[1]{\|#1\|_2}
\providecommand{\trans}{^\mathrm{T}}
\providecommand{\step}{\mathbf{1}}
\providecommand{\bq}{\mathbf{q}}
\providecommand{\bw}{\mathbf{w}}
\providecommand{\bteta}{\boldsymbol{\theta}}
\providecommand{\bx}{\mathbf{x}}
\providecommand{\bt}{\mathbf{t}}
\providecommand{\bg}{\mathbf{g}}
\providecommand{\by}{\mathbf{y}}
\providecommand{\bX}{\mathbf{X}}
\providecommand{\bbeta}{\boldsymbol{\beta}}
\providecommand{\bTeta}{\boldsymbol{\Theta}}
\providecommand{\bsigma}{\boldsymbol{\sigma}}
\providecommand{\bbR}{\mathbb{R}}
\providecommand{\setQ}{\mathcal{Q}}
\providecommand{\setW}{\mathcal{W}}
\providecommand{\setG}{\mathcal{G}}
\providecommand{\setI}{\mathcal{I}}
\providecommand{\loss}{\ell}
\providecommand{\Risk}{\mathcal{R}}
\providecommand{\obj}{\mathrm{obj}}
\providecommand{\bobj}{\mathbf{obj}}
\providecommand{\gb}{\mathrm{gb}}
\providecommand{\gs}{\mathrm{gs}}
\providecommand{\log}{\mathrm{log}}
\providecommand{\sqr}{\mathrm{sqr}}
\providecommand{\E}{\mathrm{E}}
\providecommand{\np}{k_{\mathrm{pr}}}
\providecommand{\ns}{k_{\mathrm{sp}}}
\providecommand{\halfspace}{\hspace{0.1em}}
\providecommand{\thickhline}{\noalign{\hrule height 0.25mm}}

\providecommand{\extratiny}[1]{\scalebox{0.8}{#1}}
\providecommand{\extratinytiny}[1]{\scalebox{0.72}{#1}}

\providecommand{\given}{\,|\,}
\providecommand{\meduplo}[3]{\extratinytiny{#3} & #1 & \extratinytiny{#2} }

\providecommand{\sign}{\mathrm{sign}}
\providecommand{\signg}{\zeta}

\begin{figure}[ht]
    \centering
    \includegraphics[width=\linewidth]{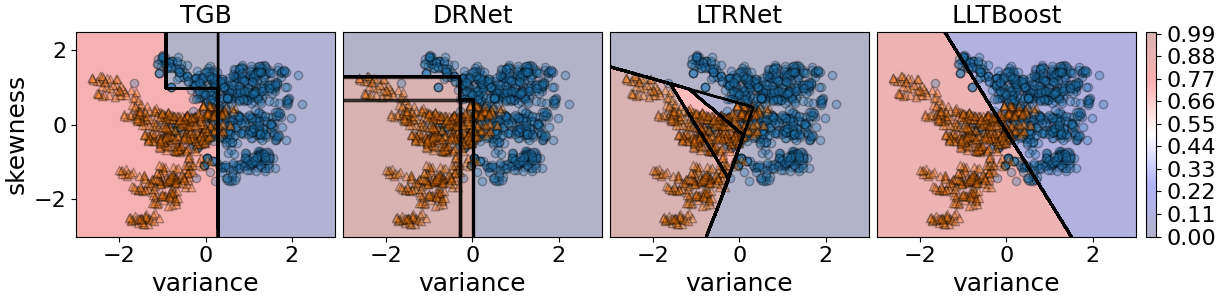}
    \caption{Visualization of the decision regions by generated rules for the Banknote dataset using LLTBoost, and baseline methods: LTRNet, TGB, and DRNet, as presented in \Cref{table:rules}. For simplicity, the models were trained using only two input variables. LLTBoost and LTRNet provide linear transformation learning of input variables resulting in oblique decision boundaries, while TGB and DRNet generate axis-parallel regions. Additionally, LTRNet and DRNet rely on backpropagation-based optimization, whereas TGB and LLTBoost use gradient boosting.}
    \label{fig:rules_plots}
\end{figure}

\providecommand{\halfspace}{\hspace{0.1em}}
\providecommand{\thickhline}{\noalign{\hrule height 0.25mm}}

\providecommand{\inner}[2]{#1\trans#2} %{\langle #1, #2 \rangle} %{#1\trans #2}
\providecommand{\with}{\!\mid\,}
\providecommand{\zeronorm}[1]{\|#1\|_0}
\providecommand{\onenorm}[1]{\|#1\|_1}
\providecommand{\twonorm}[1]{\|#1\|_2}
\providecommand{\trans}{^\mathrm{T}}
\providecommand{\step}{\mathbf{1}}
\providecommand{\bq}{\mathbf{q}}
\providecommand{\bw}{\mathbf{w}}
\providecommand{\bteta}{\boldsymbol{\theta}}
\providecommand{\bx}{\mathbf{x}}
\providecommand{\bt}{\mathbf{t}}
\providecommand{\bg}{\mathbf{g}}
\providecommand{\by}{\mathbf{y}}
\providecommand{\bX}{\mathbf{X}}
\providecommand{\bbeta}{\boldsymbol{\beta}}
\providecommand{\bTeta}{\boldsymbol{\Theta}}
\providecommand{\bsigma}{\boldsymbol{\sigma}}
\providecommand{\bbR}{\mathbb{R}}
\providecommand{\setQ}{\mathcal{Q}}
\providecommand{\setW}{\mathcal{W}}
\providecommand{\setG}{\mathcal{G}}
\providecommand{\setI}{\mathcal{I}}
\providecommand{\loss}{\ell}
\providecommand{\Risk}{\mathcal{R}}
\providecommand{\obj}{\mathrm{obj}}
\providecommand{\bobj}{\mathbf{obj}}
\providecommand{\gb}{\mathrm{gb}}
\providecommand{\gs}{\mathrm{gs}}
\providecommand{\log}{\mathrm{log}}
\providecommand{\sqr}{\mathrm{sqr}}
\providecommand{\E}{\mathrm{E}}
\providecommand{\np}{k_{\mathrm{pr}}}
\providecommand{\ns}{k_{\mathrm{sp}}}
\providecommand{\halfspace}{\hspace{0.1em}}
\providecommand{\thickhline}{\noalign{\hrule height 0.25mm}}

\providecommand{\extratiny}[1]{\scalebox{0.8}{#1}}
\providecommand{\extratinytiny}[1]{\scalebox{0.72}{#1}}

\providecommand{\given}{\,|\,}
\providecommand{\meduplo}[3]{\extratinytiny{#3} & #1 & \extratinytiny{#2} }

\providecommand{\sign}{\mathrm{sign}}
\providecommand{\signg}{\zeta}
\begin{table}[h!t]
\centering
% \scriptsize
    \caption{Example Rules for the Banknote Dataset. For simplicity, only two input variables—variance (var) and skewness (skew)—out of the four available features have been used to train each model and generate the rules. These variables represent extracted features from Wavelet Transformed images of genuine and forged banknote-like specimens.}\label{table:rules}
    \begin{subtable}[t]{\textwidth}
    \caption{\textbf{TGB}}\label{table:tgb_rules}
    \begin{tabular}{p{0.07\linewidth}p{0.05\linewidth}p{0.85\linewidth}}
    \thickhline
        +1.46 & if & True \\
        -4.58  & if & var $\geq$ 0.295\\
        -12.7 & if & skew $\geq$ 0.975 \textbf{\&} var$\leq$ 0.295 \textbf{\&} var $\geq$ -0.918
    \end{tabular}
    \end{subtable}
    \begin{subtable}[t]{\textwidth}
    \caption{\textbf{DRNet}}\label{table:drnet_rules}
    \begin{tabular}{p{0.07\linewidth}p{0.05\linewidth}p{0.85\linewidth}}
    \thickhline
        True & if & var $\leq$ 0.022 \textbf{\&} skew $\leq$ 0.67\\
        True  & if & var $\leq$ -0.29 \textbf{\&} skew $\leq$ 1.28\\
        True & if & var $\leq$ -0.29 \textbf{\&} skew $\leq$ 0.67 \\
        False & if & else
    \end{tabular}
    \end{subtable}
    \begin{subtable}[t]{\textwidth}
    \caption{\textbf{LTRNet}}\label{table:ltrnet_rules}
    \begin{tabular}{p{0.07\linewidth}p{0.05\linewidth}p{0.85\linewidth}} 
    \thickhline
        -25.9 & if & True \\ 
        11.94 & if & -11.16\halfspace var +3.88\halfspace skew $\geq$ -4.1 \textbf{\&}
        -3.36\halfspace var -10.3 \halfspace skew $\geq$ -5.95 \\
        3.45 & if & -2.4\halfspace var -1.28\halfspace skew $\geq$ 2.45\\
        -4 & if & 2.29\halfspace var +2.28\halfspace skew $\geq$ -0.99  \textbf{\&}
        2.22\halfspace var +2.26\halfspace skew $\geq$ -0.44 \\
        3.45 & if & -2.04\halfspace var -6.11\halfspace skew $\geq$ -3.51\\
        11.42 & if & -3.8\halfspace var -11.67\halfspace skew $\geq$ -6.73 \textbf{\&}
        -13.95\halfspace var +4.77\halfspace skew $\geq$ -3.58 \\
        -7.31 & if & 6.35\halfspace var -1.93\halfspace skew $\geq$ 1.14
    \end{tabular}
    \end{subtable}
    \begin{subtable}[t]{\textwidth}
    \caption{\textbf{LLTBoost}}\label{table:lltb_rules}
    \begin{tabular}
    {p{0.07\linewidth}p{0.05\linewidth}p{0.85\linewidth}} 
    \thickhline
        +1.79 & if & True \\
        -3.85 & if & 2.81\halfspace var +1.5\halfspace skew $\geq$ -0.28 \\
        % \thickhline
    % \normalize
    \end{tabular}
    \end{subtable}
\end{table}

\section{Introduction}\label{sec:intro}

\providecommand{\inner}[2]{#1\trans#2} %{\langle #1, #2 \rangle} %{#1\trans #2}
\providecommand{\with}{\!\mid\,}
\providecommand{\zeronorm}[1]{\|#1\|_0}
\providecommand{\onenorm}[1]{\|#1\|_1}
\providecommand{\twonorm}[1]{\|#1\|_2}
\providecommand{\trans}{^\mathrm{T}}
\providecommand{\step}{\mathbf{1}}
\providecommand{\bq}{\mathbf{q}}
\providecommand{\bw}{\mathbf{w}}
\providecommand{\bteta}{\boldsymbol{\theta}}
\providecommand{\bx}{\mathbf{x}}
\providecommand{\bt}{\mathbf{t}}
\providecommand{\bg}{\mathbf{g}}
\providecommand{\by}{\mathbf{y}}
\providecommand{\bX}{\mathbf{X}}
\providecommand{\bbeta}{\boldsymbol{\beta}}
\providecommand{\bTeta}{\boldsymbol{\Theta}}
\providecommand{\bsigma}{\boldsymbol{\sigma}}
\providecommand{\bbR}{\mathbb{R}}
\providecommand{\setQ}{\mathcal{Q}}
\providecommand{\setW}{\mathcal{W}}
\providecommand{\setG}{\mathcal{G}}
\providecommand{\setI}{\mathcal{I}}
\providecommand{\loss}{\ell}
\providecommand{\Risk}{\mathcal{R}}
\providecommand{\obj}{\mathrm{obj}}
\providecommand{\bobj}{\mathbf{obj}}
\providecommand{\gb}{\mathrm{gb}}
\providecommand{\gs}{\mathrm{gs}}
\providecommand{\log}{\mathrm{log}}
\providecommand{\sqr}{\mathrm{sqr}}
\providecommand{\E}{\mathrm{E}}
\providecommand{\np}{k_{\mathrm{pr}}}
\providecommand{\ns}{k_{\mathrm{sp}}}
\providecommand{\halfspace}{\hspace{0.1em}}
\providecommand{\thickhline}{\noalign{\hrule height 0.25mm}}

\providecommand{\extratiny}[1]{\scalebox{0.8}{#1}}
\providecommand{\extratinytiny}[1]{\scalebox{0.72}{#1}}

\providecommand{\given}{\,|\,}
\providecommand{\meduplo}[3]{\extratinytiny{#3} & #1 & \extratinytiny{#2} }

\providecommand{\sign}{\mathrm{sign}}
\providecommand{\signg}{\zeta}

Additive ensembles of symbolic rules or rule sets~\cite{lakkaraju2016interpretable,wei2019generalized,friedman2008predictive,furnkranz1999separate} are an interpretable and often accurate type of prediction model where the prediction for an input $\bx$ is given as $\mu(f(\bx))$ with an activation function $\mu: \bbR \to \bbR$ and $f: \bbR^d \to \bbR$ being a weighted sum
\begin{align}\label{eq:fx}
    f(\bx) = \beta_0 + \sum_{i=1}^{r}\beta_i q_i(\bx) \enspace .
\end{align}
Here, the coefficients $\beta_i$ play the role of the rule consequents and the $q_i(\bx) \in \{0, 1\}$ map the input vector to binary outputs and thus correspond to logical rule conditions.
This form allows for a range of modeling problems (regression, classification, counting regression) with a probabilistic interpretation by choosing a suitable activation function such that $\mathbb{E}[Y \mid X= \bx] =\mu(f(\bx))$.
Further, same as generalized linear models~\cite{mccullagh2019generalized} and generalized additive models~\cite{hastie2017generalized}, additive rule ensembles are modular~\cite{murdoch2019definitions}, meaning that due to the simple rule aggregation as a sum, interpretation of the ensemble can be broken down into interpretation of individual rules.
Therefore, the overall model is principally simulatable~\cite{murdoch2019definitions} by a human interpreter if the individual rules are.
Traditionally this is true, because rule conditions are given as conjunctions of simple threshold functions on individual input variables. Formally, for rule $i$ there are thresholds $t_{i, j}$, input variables $u_{i,j} \in \{1, \dots, d\}$, and indicators $s_{i,j} \in \{-1, 1\}$ for $j \in \{1, \dots k_r\}$ such that
\begin{align}
    q_i(\bx) &= \prod_{j=1}^{k_r} p_{i,j}(\bx),  \label{eq:qi}\\
    p_{i,j}(\bx) &= \step \{s_{i,j}x_{u_{i,j}}\geq t_{i,j}\} \enspace .\label{eq:pij}
\end{align}
Geometrically, this choice means that rule conditions form axis-parallel hyper-rectangles in the input space (see the first two subfigures, TGB and DRNet in \Cref{fig:rules_plots}).
While this form ensures a high degree of interpretability for individual rules and can be learned efficiently using the gradient boosting approach~\cite{dembczynski2010ender,friedman2001greedy,boley2021better,yang2024orthogonal}, it relies on having access to a curated set of ideally independent input features so that a small ensemble of axis-parallel regions can describe the target variable well. Absent such features, reaching sufficient accuracy requires ramping up the complexity in terms of the number and size of individual rules, which ultimately diminishes the practical interpretability of the model. 

This paper introduces an extension of classical rule ensemble architecture by logical propositions with learnable sparse linear transformations on the input variables, i.e., propositions of the form $\bx\trans\bw \geq t$, allowing decision regions that correspond to general polytopes with oblique faces. For example, the LLTBoost subfigure in \Cref{fig:rules_plots} illustrates classification using a single rule with a linear form, creating an oblique decision boundary, whereas TGB requires a larger number of rules and conditions. The corresponding rule is also presented in \Cref{table:lltb_rules}.
This is reminiscent of work that has been done in the context of decision trees, in particular soft trees~\cite{irsoy2012soft,xu2022one}.
However, tree-based approaches struggle with a large number of rules (leaves) and the soft thresholds further diminish the interpretability of these models.
In contrast, we propose an architecture that obeys the traditional rule semantics of hard threshold and that can be restricted to a desired level of complexity. The proposed semantic neural architecture can be considered an adaption of discrete neural network rule learning as explored in works such as DRNet~\cite{qiao2021learning} and other variations like ENRL~\cite{shi2022explainable}, RLNet~\cite{dierckx2023rl}, and RRL~\cite{wang2023learning}. 
To train this architecture, we introduce the logistic linear transformation boosting (LLTBoost) method, a sequential greedy optimization based on gradient boosting where each proposition, $p_{i,j}$, formed as $\bx\trans\bw \geq t$, is learned by maximizing the boosting objective, which is reduced to a sequence of weighted regularized binary classification problems.
We also introduce the linear transformation rule network (LTRNet) as a baseline for LLTBoost, which is a parallel optimization of all parameters through the minimization of a continuous penalized risk function followed by a post-processing discretization step for the proposed architecture.
To better highlight the differences between each method, \Cref{fig:rules_plots} and \Cref{table:rules} present an example using the Banknote dataset. Originally containing four input variables, only two are used here for simplicity in visualization. The empirical evaluation later includes results using all four variables. As shown, LLTBoost offers a more concise and interpretable set of rules through oblique decision regions, while other methods require more conditional terms to achieve the same predictions.
As empirically demonstrated, LLTBoost significantly improves the risk/complexity trade-off. In all ten common benchmark datasets, LLTBoost produces rule ensembles with the same test risk but lower complexity than other baseline methods. Notably, in seven of these datasets, the complexity is reduced by a factor of more than two. Furthermore, the proposed method requires computation time in the same order of magnitude as the TGB algorithm.
Importantly, the complexity measurement employed here does take into account the additional complexity introduced by the linear transformation.

\begin{figure}[h!]
    \centering
    \includegraphics[width=1\linewidth]{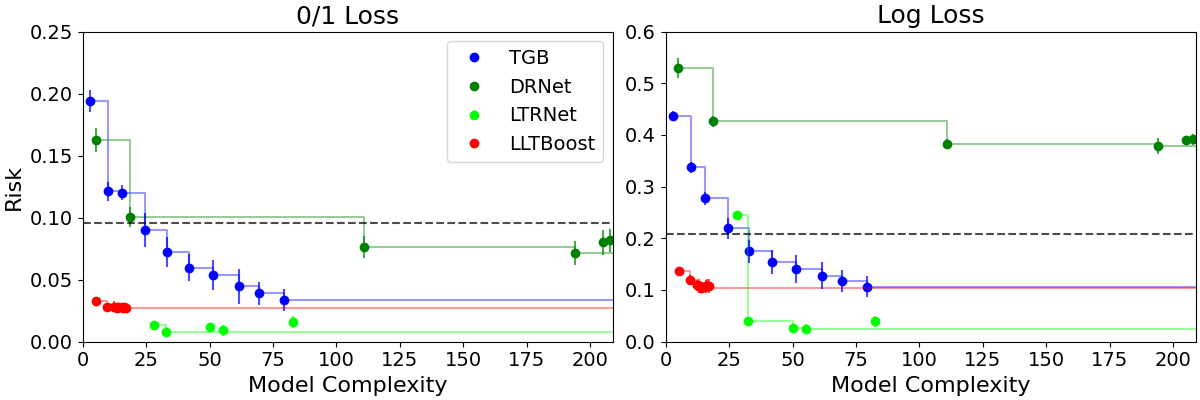}
    \caption{Risk-Complexity trends for Banknote dataset, comparing LLTBoost, with baseline methods: LTRNet, TGB and DRNet. Each point represents the mean risk and mean complexity over ten repetitions, corresponding to the method's parameter, such as the maximum number of rules for TGB and LLTBoost, and the penalty term value for DRNet and LTRNet. The black dashed line indicates the test risk targets for selecting the ensembles of each method, as represented in \Cref{table:complexity_table}.}
    \label{fig:risk_complexity_trend_banknote}
\end{figure}
\section{Background}\label{sec:background}
In this section, we provide background information on gradient boosting and weighted logistic regression which form the basis of our proposed method, LLTBoost. 

\paragraph{\textbf{Risk Minimization}}
We consider a standard supervised learning setting where we are given a training set $\{(\bx_i, y_i): i=1, \dots, n\}$, $\bx_i = (x_{i,1},\dots,x_{i,d})$ sampled iid from the joint distribution of an input variable $X$ and an output variable $Y$, and aim to fit a prediction model $\mu(f(\bx))$ that minimizes the prediction risk $\E[\loss(Y, \mu(f(X)))]$ by minimizing the empirical risk, i.e., to find
\begin{equation}
    f^* = \underset{f}{\argmin} \left( \frac{1}{n}\sum_{i=1}^n \loss(y_i,\mu(f(\bx_i))) + \Omega(f) \right) \enspace .
\end{equation}
Here, $\mu$, $\Omega$ and $\loss$ are an activation, a regularization function and a loss function specific to the learning problem.
A traditional non-probabilistic choice for the case of binary classification where $y \in \{0, 1\}$ is to use the step function $\mu(f(\bx)) = \step\{f(\bx)\}$ together with the 0/1 loss  $\loss_{01}(y, y')= \step\{y\neq y'\}$.
For probabilistic models that assume that $Y \given X=\bx$ is some family of distributions parameterized by $\E[Y \given X=\bx] = \mu(f(\bx))$, a general approach is to consider the loss function derived as deviance~\cite{mccullagh2019generalized}, i.e.,
$\loss_\mathrm{log}(y, y')=\log P(y \given \mu^{-1}(y)) - \log P(y \given y')$.
For regression where $y \in \bbR$ and $Y \given X=\bx$ is assumed to be normally distributed with mean $\mu(a) = a$ results in the squared loss $\loss_{\sqr}(y,y') = (y-y')^2/2$.
For classification with $\mu(a) = 1/(1+\exp(-a))$ this results in the logistic loss defined as $\loss_{\log}(y, y') = -y\log y' - (1-y)\log(1-y')$.

\paragraph{\textbf{Sparse Logistic Regression}}
For linear $f(\bx; \bw, t)=\inner{\bx}{\bw}+t$, which we will utilize in this work as a building block, the above approach is known as logistic regression.
Specifically, we will be interested in minimizing the $L_1$ regularized empirical risk
\begin{equation}\label{eq:reglogregmin}
    \bw^*, t^* = \argmin \sum_{i=1}^n \loss_\mathrm{log} (y_i, f(
    \bx_i;\bw,t)) + \lambda_s\onenorm{\bw}
\end{equation}
where $\lambda_s = \mathrm{min} \{ \lambda \in \bbR^+ : \zeronorm{\bw} = s\}$ is chosen minimally to yield an $s$-sparse solution.

\paragraph{\textbf{Rules and Complexity}} For additive rule ensembles, the function $f$ is given in the form of \Cref{eq:fx}.
In this case, the objective is not only to find an $f$ with small risk but also one that is comprehensible, i.e., has a small number of ideally short rules.
Comprehensibility can be measured on various levels of granularity. A natural measure of model complexity is given by
\begin{equation}
    C(f) = r + \sum_{i=1}^{r} c(q_i) \label{eq:c} 
\end{equation}
where $r$ is the number of conjunctions (rules) and $c(q_i)$ denotes the complexity of each conjunction, given by the total number of variables and thresholds as $c(q_i) = 2k_{i}$. Similar notions of complexity have been used by Yang et al. \cite{yang2024orthogonal} and Qiao et al. \cite{qiao2021learning}. Later, we introduce a more generalized definition of $c(q_i)$.
%%%%%%%%%%%%%%%%%%%%%%%%%%%%%%%%%%%%%%%%%%%%%%%%%%%%%%%
%%%%%%%%%%%%%%%%%%%%%%%%%%%%%%%%%%%%%%%%%%%%%%%%%%%%%%%
\paragraph{\textbf{Gradient Boosting}}
In rule learning, a sequence of rules can be produced through gradient boosting. Specifically, the fully corrective boosting variant~\cite{shalev2010trading}, starts with an empty rule with a consequent of $\beta_0$ which $\beta_0 = \argmin(\sum_{i=1}^n \loss(y_i, \step(\beta)))$ and the final prediction function is obtained iteratively for $m = 1,\dots, r$ as 
\begin{align}
    f^{(m)}(\bx) &= \beta_0^{(m)} + \beta_{1}^{(m)}q_1(\bx)+\dots+\beta_{m}^{(m)}q_m(\bx) \label{eq:ruleboosting} \\
    q_m &= \argmax \{\obj(q;f^{(m-1)}) \with q \in \setQ \} \\
     \bbeta^{(m)} &= \underset{\bbeta \in \bbR^{m+1}}{\argmin} \left(\sum_{i=1}^n \loss(y_i,\inner{(1, q_1(\bx_i),\dots, q_{m}(\bx_i))}{\bbeta})\right) \label{eq:beta} \enspace .
\end{align}
$\bbeta^{(m)} = (\beta_0^{(m)}, \dots, \beta_m^{(m)})$ denotes that at each iteration of boosting, rule weights are refitted. The role of the objective function $\obj: \bbR^n \to \bbR$ is to approximate the empirical risk reduction achieved by adding a candidate rule.
Here, we consider the gradient sum objective function~\cite{shalev2010trading,yang2024orthogonal}
\begin{align}
    \obj(\bq) &= |\bg\trans \bq |
    \label{eq:obj_gb} \\
    \bg &= (g(\bx_1),\dots, g(\bx_n)) \label{eq:g}\\
    g(\bx_i) &= \frac{\partial \loss(y_i, f^{(m-1)}(\bx_i))}{\partial f^{(m-1)}(\bx_i)}, \; i=1,\dots, n  \label{eq:g_i} \\
    \bq &= (q(\bx_1),\dots, q(\bx_n)) \enspace .
\end{align}
%%%%%%%%%%%%%%%%%%%%%%%%%%%%%%%%%%%%%%%%%%%%%%%%%%%%%%%
%%%%%%%%%%%%%%%%%%%%%%%%%%%%%%%%%%%%%%%%%%%%%%%%%%%%%%%

\section{Sparse Linear Representation Rule Learning}\label{sec:methods}
In this section, we propose a semantic architecture to represent an additive rule ensembles model and a method for learning a sparse linear representation of the input vector, $\bx$, for each proposition and output. We aim to extend the proposition term defined in \Cref{eq:pij} by introducing a linear transformation of the input variables, bounded by a threshold, rather than relying on individual variables alone. The key idea to achieve this goal is to redefine the propositions as:
\begin{align}
    p_{i,j}(\bx) = \step{\{\bx\trans\bw_{i,j} \geq t_{i,j}\}} \label{eq:linearproposition}
\end{align}
where for the $j$th proposition of $i$th rule, $\bw_{i,j} \in \bbR^d$ is the learnable weight vector for a linear transformation of the input vector and $t_{i,j} \in \bbR$ is a learnable threshold. This linear transformation enables the partitioning of the data space with oblique faces, as opposed to axis-parallel faces in classic rule ensembles.
Based on our new definition of propositions constituted of a linear transformation of input variables, we need to measure the complexity of each linear transformation. By maintaining the generality of model complexity definition \Cref{eq:c}, we compute the complexity of each conjunction, $c(q_i)$ in \Cref{eq:cq}, by counting the number of non-zero weights of each linear transformation in addition to the number of rules and thresholds within each rule.
\begin{align}
    c(q_i) &= k_i + \sum_{j=1}^{k_{i}}\zeronorm{\bw_{i,j}} \quad , \quad i=1,\dots,r.   \label{eq:cq}
\end{align}
\begin{figure}[ht]
    \centering
    \includegraphics[width=0.9\linewidth]{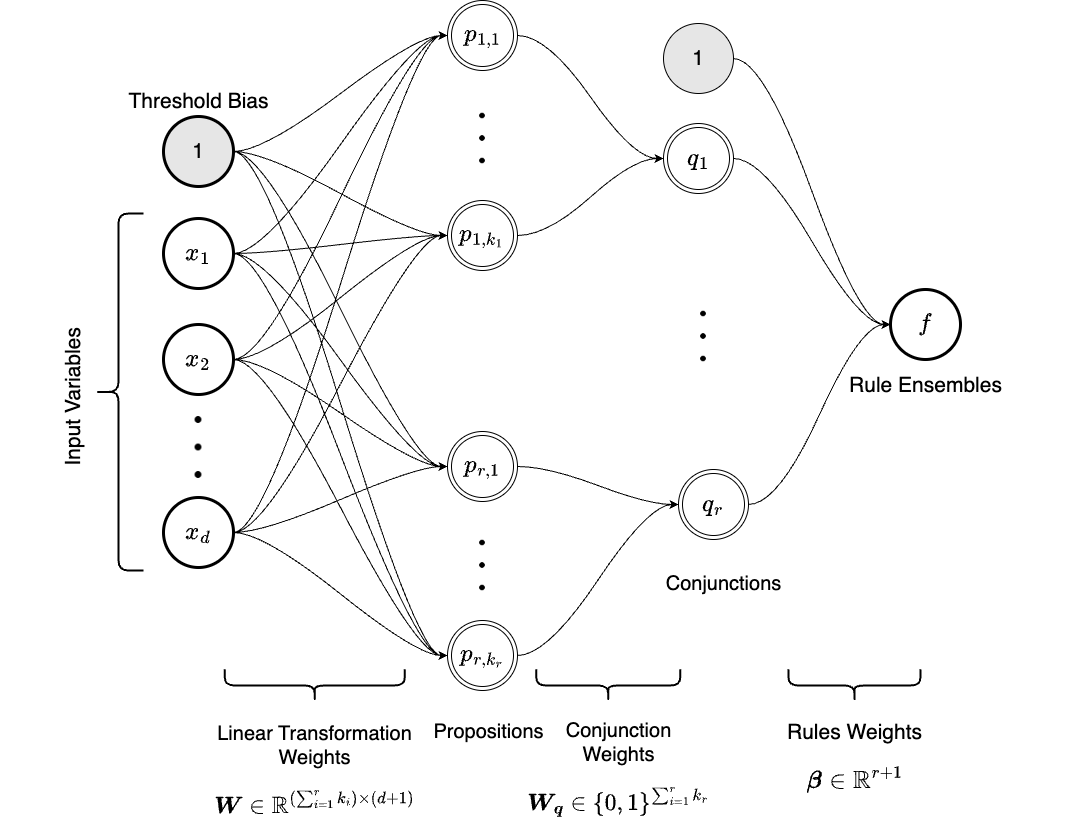}
    % \vspace{.3in}
    \caption{Basic architecture for linear representation learning for additive rule ensembles.}
    \label{fig:architecture}
\end{figure}
\subsection{Basic Architecture}

\Cref{fig:architecture} shows the architecture that integrates linear representation learning into classic additive rule ensembles. This structure can be viewed as a three-layer model, where each layer corresponds to the core components of a rule ensemble model, as discussed in \Cref{sec:background} and \Cref{eq:linearproposition}. Below, we outline the role of each layer in the learning process.

\paragraph{\textbf{Propositions}}
In this layer of the architecture shown in \Cref{fig:architecture}, propositions, denoted as $\{p_{i,j}(\bx)\with i=(1,\dots,r), j=(1,\dots,k_i)\}$, are formed through a linear transformation of the input feature variables $\bx$ (of size $d$), along with a bias term acting as a lower threshold as we defined in \Cref{eq:linearproposition}. This combination maps the original input space into a new feature space. 
\paragraph{\textbf{Conjunctions}}
This layer is responsible for learning the conjunctions defined in \Cref{eq:qi} by combining the relevant propositions. Each node acts as an ``AND'' gate, receiving boolean inputs from the propositions. Therefore, the conjunction weights in this layer should be learned as binary values. Conjunctions do not share any common proposition nodes.
\paragraph{\textbf{Rule Weight}} In the last layer, the rules weights vector, $\bbeta$ is learned to form the final prediction function defined in \Cref{eq:beta}.
%%%%%%%%%%%%%%%%%%%%%%%%%%%%%%%%%%%%%%%%%%%%%%%%%%%%%%%
%%%%%%%%%%%%%%%%%%%%%%%%%%%%%%%%%%%%%%%%%%%%%%%%%%%%%%%
\subsection{Logistic Linear Transformation Boosting} \label{sec:fs}
In this section, we develop a base-learner for gradient boosting of additive rule ensembles with the proposed architecture. For this, we follows the common approach of greedily optimizing one proposition at a time. This means finding one transformation $\bw_{m,j}$ and $t_{m,j}$ at a time that, together with already fixed propositions $p_{m,1}, 
\dots, p_{m, j-1}$, maximize the boosting objective~\eqref{eq:obj_gb}. 
As we will show below, a key insight is that maximizing the objective function for an individual linear transformation can be reduced to minimizing the weighted 0/1-loss for certain binary classification problems.
Finally, as our goal is to minimize the overall complexity, we further break down the optimization into testing one sparsity level $\|\bw_{m,j}\|=s$ at a time and only accepting a decrease in sparsity, from $s$ to $s+1$, if the corresponding risk reduction appears significant based on a held-out validation set. 

In more detail, we firstly break down the absolute value in the objective function defined in \Cref{eq:obj_gb} into positive and negative parts, $\obj_+(\bq)=\inner{\bg}{\bq}$ and $\obj_-(\bq)=\inner{-\bg}{\bq}$, and write 
\begin{equation}
\max |\inner{\bg}{\bq}|=\max \{\inner{\bg}{\bq}, \inner{-\bg}{\bq} \! : \, \bq \in \setQ\} \enspace .
\end{equation}
Then, given the set of still selectable training examples $\setI_{j}=\{i \with p_{m, 1}(x_i)=\dots=p_{m, j-1}(x_i)=1\}$, we rewrite the objective for $\signg \in \{-1, 1\}$ as follows:
\begin{align}
    \obj^{(\signg)}_{j}(\bw, t) &= \sum_{i\in \setI_j} \signg g_i \step\{\bx_i\trans\bw \geq t\} \\
    &=\sum_{i\in \setI^+_j} \signg g_i - \sum_{i\in \setI^+_j} \signg g_i\step\{\bx_i\trans\bw < t\} + \sum_{i\in \setI^-_j} \signg g_i\step\{\bx_i\trans\bw \geq t\}\\
    &= \sum_{i\in \setI^+_j} \signg g_i - \underbrace{\sum_{i \in \setI_j} |g_i| \loss_{01}(\signg \mathrm{sign}(g_i), \step\{\signg \bx_i\trans\bw\geq t\})}_{\Risk^{(\signg)}_{01}(\bw, t)}
\end{align}
where we used the symbols $\setI^+_j = \{i \in \setI_j \with \signg g_i \geq 0 \}$ and $\setI^-_j = \{i \in \setI_j \with \signg g_i < 0 \}$.

Thus, maximizing the objective function for proposition $j$ is equivalent to minimizing the weighted sum of 0/1-losses when interpreting the weight vector as a linear separating hyperplane for predicting the gradient signs.
This allows us to utilize linear classification algorithms to optimize the weight parameters.

Before we do that, we incorporate the idea of controlling the sparsity of the weights into the base-learner. 
For that, we require a validation set, and sequentially test the optimal weights for sparsity levels $s=1$ up to some maximum number of non-zero element $k_s$, only accepting an $(s+1)$-sparse solution if it significantly improves the validation risk over the optimal $s$-sparse solution.
To find the optimal $\bw^*_{m,j}$ and $t^*_{m,j}$ for some sparsity level $s$ we can find
\begin{equation}
    \Risk^{(\signg)}_{01}(\bw^*_{m,j},t^*_{m,j}) = \min \{\Risk^{(\signg)}_{01}(\bw,t) : t \in \bbR, \bw\in \bbR^d, \zeronorm{\bw}=s\} \enspace .
\end{equation}
Since this is a hard optimization problem, it is commonly optimized using a convex surrogate function. In this work, we propose to optimize the $L_1$-regularized risk with respect to the logistic loss, $\loss_{\log}$
\begin{equation}\label{eq:l1logreg}
    \bw^*_{m,j},t^*_{m,j} = \argmin \Risk^{(\signg)}_{\log}(\bw,t; |\bg|) + \lambda_s\onenorm{\bw}
\end{equation}
Similarly, as discussed in \Cref{eq:reglogregmin}, $\lambda_s$ is the smallest regularization value that results in a solution with exactly $s$ non-zero weights. This value can be found implicitly by a binary search strategy through solving the convex optimization problem of \Cref{eq:l1logreg}. We close this section with the pseudocode \Cref{alg:FS_learner} that presents all details of the developed base learner.

%%%%%%%%%%%%%%%%%%%%%%%%%%%%%%%%%%%%%%%%%%%%%%%%%%%%%%%
%%%%%%%%%%%%%%%%%%%%%%%%%%%%%%%%%%%%%%%%%%%%%%%%%%%%%%%
\subsection{Linear Transformation Rule Network}\label{sec:fj}
\begin{figure}[h]
    \centering
    \begin{subfigure}{0.49\textwidth} 
        \includegraphics[width=\linewidth]{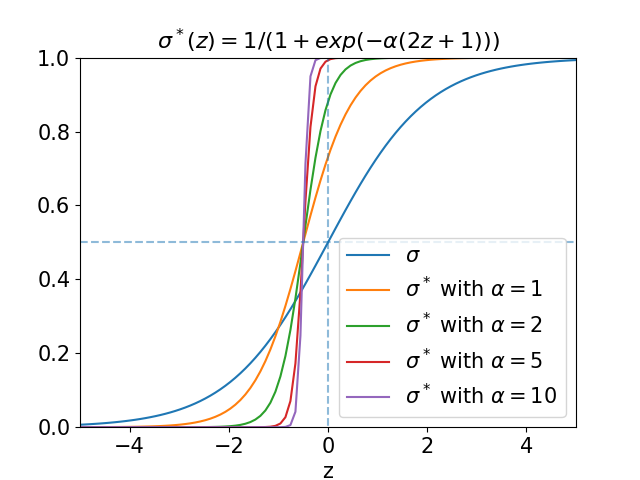}
        \caption{Modified Sigmoid Function}
        \label{fig:modifiedsigmoid}
    \end{subfigure}
    \begin{subfigure}{0.49\textwidth} 
        \includegraphics[width=\linewidth]{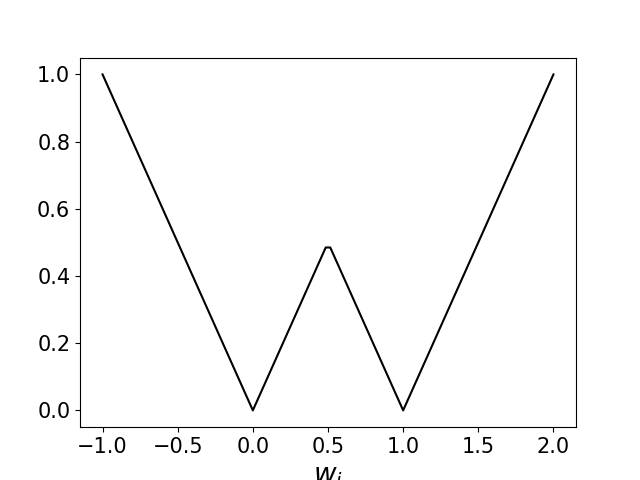}
        \caption{W-shaped Penalty}
        \label{fig:wpenalty}
    \end{subfigure}
    \caption{Functions employed by LTRNet method to train the discrete rule ensembles network.}
\end{figure}

To ensure a consistent and precise evaluation, we also provide LTRNet as a parallel rule learning method based on backpropagation optimization as a baseline for LLTBoost.
In this method, the full architecture illustrated in \Cref{fig:architecture} is assumed to learn all parameters simultaneously through gradient descent optimization (backpropagation). The output of each proposition as a conditional term and conjunction as an ``AND'' gate is a boolean or binary value, then ideally, we can assume a step activation function for propositions and conjunction nodes. However, due to the gradient vanishing problem inherent in the step function, differentiable activation functions that approximate step function behaviour must be used to enable backpropagation. In the propositions layer, a sigmoid function $\sigma(z) = 1/(1+e^{-z})$, is used for the conditional terms of propositions. However, in the conjunctions layer, simply replacing the step function with a sigmoid is insufficient because we need to model an ``AND'' gate, which involves a linear combination of binary inputs from the proposition units—essentially mimicking the product of binary values in a rule ensemble model. To address this, we introduce the concepts of a \textit{balancing bias} and a \textit{modified sigmoid} function to maintain the logical output of the conjunction units. For each conjunction unit, the balancing bias is represented as the negative norm of the conjunction weights $-||\bold{w_q}_k||, k \in \{1,\dots,r\}$. The modified sigmoid function is defined as $\sigma^*(z) = 1/(1+e^{-\alpha(2z+1)})$ where $\alpha$ is a hyperparameter that controls the steepness of the sigmoid, allowing it to better approximate the behaviour of a step function. \Cref{fig:modifiedsigmoid} compares $\sigma^*(z)$ for different values of $\alpha$ against the standard sigmoid, $\sigma(z)$. Together, the balancing bias and modified sigmoid effectively help with simulating an ``AND'' gate for a linear combination of binary values through backpropagation.
Moreover, the weights of the conjunction layer should be binary to determine how many propositions contribute to forming a conjunction node or rule body. One approach to achieve binarized conjunction weights is to apply a specific penalty term that drives these weights toward either 0 or 1~\cite{bai2018proxquant}. Similar to the $L_1$ norm regularization, which shrinks parameters to 0, we propose using a W-shaped penalty term displayed in \cref{fig:wpenalty} as a piecewise-componentwise function that narrows the conjunction weights to 0 or 1. In addition to the W-shaped penalty, we enforce sparsity in both the linear transformation of input variables and the number of rules by applying an $L_1$ norm penalty to the propositions layer and rule weights.
\paragraph{\textbf{Discretization Step}}
Considering the continuous optimization approach, for enhanced interpretability, the fully-trained architecture must be transformed into a discrete structure. This post-processing phase includes the following essential steps:
\begin{itemize}
    \item Rounding the conjunction layer weights to either zero or one, based on the nearest value.
    \item Replacing the modified sigmoid function with a step function, using the same threshold.
    \item Replacing the sigmoid function in the propositions layer with a step function.
    \item Rounding the linear transformation weight vectors using a threshold determined by the maximum magnitude within the vector.
\end{itemize}
LTRNet can be an extention upon DRNet~\cite{qiao2021learning} which is neural network-based rule learning method representing a two-layer architecture that maps and aggregates the set of If-Then rules for binary classification. LTRNet incorporates an additional linear transformation layer and weighted rules. In contrast to DRNet, LTRNet does not rely on binarized pre-proccessed inputs and can learn any linear combination of input variables.

\newcommand{\bp}{\mathbf{p}}
\newcommand{\vali}{\mathrm{val}}
\begin{algorithm}[h!]
\SetAlgoNlRelativeSize{-1}
\SetKwFor{For}{for}{do}{}
\SetKwIF{If}{ElseIf}{Else}{if}{then}{else if}{else}{}
% \SetAlgoLined
\DontPrintSemicolon
\caption{LLTBoost Base Learner}\label{alg:FS_learner}
\textbf{Input:} training and validation data $(\bx_i)_{i=0}^n$, $\bg = (g(\bx_i))_{i=1}^n$, $(\bx^{\mathrm{val}}_i,y^{\mathrm{val}}_i)_{i=1}^{n^{\prime}} $,\;\qquad maximum number of propositions $k_{\mathrm{r}}$ and non-zero weights $k_{\mathrm{s}}$,\;
\qquad validation risk tolerance $\epsilon_s$, and objective tolerance $\epsilon_p$.\;
$q_{p-1} \gets \emptyset$ \;
$\obj_{p-1} \gets \emptyset$ \;
\For{\textnormal{\textbf{each}} $\signg \in \{1, -1\}$}{
\For{$p = 1$ \textbf{to} $k_r$}{
    $\setI_{p-1} = \{i \with q_{p-1}(\bx_i) = 1, i=1,\dots ,n\}$\;
    $\obj_{p-1} = |\inner{\signg\bg}{q_{p-1}(\bx_i)_{i=1}^n}|$ \;
    $R_{p} = \infty$ \;
    \For{$s = 1 $ \textbf{to} $k_{\mathrm{s}}$}{
        $\bw_s, t_s \gets \text{L1LogReg}\left( \{(\bx_i, \step\{\signg g(\bx_i)\geq 0\}, |g_i|)\}_{i\in \setI_{p-1}}, s \right)$\;
        $p_s, \, q_s \gets \step\{\inner{(\cdot)}{\bw_s} \geq t_s\}, \quad q_{s-1}(\cdot) p_s(\cdot)$\;
        $R_s = \sum_{i=1}^{n^{\prime}} \loss(y_i^{\mathrm{val}},\mu(f^{(m)}(\bx_i^{\mathrm{val}})))$\;
        \If{$(R_s-R_p)/R_p \leq \epsilon_s$}{
            $q_p, R_p \gets q_s,  R_s$ \;
            }
    }
    $\obj_p = |\signg\bg\trans{q_p(\bx_i)_{i=1}^n}|$\;
    \If{$(\obj_p-\obj_{p-1})/\obj_{p-1} \leq \epsilon_p$}{
        $q^{(\signg)}, \obj^{(\signg)} \gets q_{p-1}, \obj_{p-1}$\;
        \textbf{break inner loop} \;
    }
}
}
\textbf{Output} $q^{(1)}$ \textbf{if} $\obj^{(1)} > \obj^{(-1)}$ \textbf{else} $q^{(-1)}$\;
\end{algorithm}

\providecommand{\inner}[2]{#1\trans#2} %{\langle #1, #2 \rangle} %{#1\trans #2}
\providecommand{\with}{\!\mid\,}
\providecommand{\zeronorm}[1]{\|#1\|_0}
\providecommand{\onenorm}[1]{\|#1\|_1}
\providecommand{\twonorm}[1]{\|#1\|_2}
\providecommand{\trans}{^\mathrm{T}}
\providecommand{\step}{\mathbf{1}}
\providecommand{\bq}{\mathbf{q}}
\providecommand{\bw}{\mathbf{w}}
\providecommand{\bteta}{\boldsymbol{\theta}}
\providecommand{\bx}{\mathbf{x}}
\providecommand{\bt}{\mathbf{t}}
\providecommand{\bg}{\mathbf{g}}
\providecommand{\by}{\mathbf{y}}
\providecommand{\bX}{\mathbf{X}}
\providecommand{\bbeta}{\boldsymbol{\beta}}
\providecommand{\bTeta}{\boldsymbol{\Theta}}
\providecommand{\bsigma}{\boldsymbol{\sigma}}
\providecommand{\bbR}{\mathbb{R}}
\providecommand{\setQ}{\mathcal{Q}}
\providecommand{\setW}{\mathcal{W}}
\providecommand{\setG}{\mathcal{G}}
\providecommand{\setI}{\mathcal{I}}
\providecommand{\loss}{\ell}
\providecommand{\Risk}{\mathcal{R}}
\providecommand{\obj}{\mathrm{obj}}
\providecommand{\bobj}{\mathbf{obj}}
\providecommand{\gb}{\mathrm{gb}}
\providecommand{\gs}{\mathrm{gs}}
\providecommand{\log}{\mathrm{log}}
\providecommand{\sqr}{\mathrm{sqr}}
\providecommand{\E}{\mathrm{E}}
\providecommand{\np}{k_{\mathrm{pr}}}
\providecommand{\ns}{k_{\mathrm{sp}}}
\providecommand{\halfspace}{\hspace{0.1em}}
\providecommand{\thickhline}{\noalign{\hrule height 0.25mm}}

\providecommand{\extratiny}[1]{\scalebox{0.8}{#1}}
\providecommand{\extratinytiny}[1]{\scalebox{0.72}{#1}}

\providecommand{\given}{\,|\,}
\providecommand{\meduplo}[3]{\extratinytiny{#3} & #1 & \extratinytiny{#2} }

\providecommand{\sign}{\mathrm{sign}}
\providecommand{\signg}{\zeta}
\newcolumntype{C}{>{\centering\arraybackslash}}
\newcolumntype{R}{>{\raggedright\arraybackslash}}
\newcolumntype{L}{>{\raggedleft\arraybackslash}}

\begin{table}[h!]
\centering
\caption{Model complexity comparison of LLTBoost (LLTB), LTRNet, TGB, and DRNet methods on benchmark datasets using two test risk targets: one derived from TGB (top ten rows) and another from DRNet (last five rows). The risk target ($R_T$) for each dataset is computed as the average test risk of TGB/DRNet across different complexity levels. For the TGB-based risk target, squared loss is used for regression and log loss for classification. For the DRNet-based risk target, a 0/1 loss is applied. The top five rows correspond to regression datasets, while the remaining rows represent classification datasets. Three numbers are reported for each experiment, representing the median (center), lower (left), and upper (right) bounds of the 77\% confidence interval.}\label{table:complexity_table}
\begin{tabularx}{\textwidth}{CX|Cm{1.2cm}Lm{0.6cm}Cm{0.6cm}Rm{0.6cm}Lm{0.6cm}Cm{0.6cm}Rm{0.6cm}Lm{0.6cm}Cm{0.6cm}Rm{0.6cm}Lm{0.6cm}Cm{0.6cm}Rm{0.6cm}}
\hline
Dataset & $R_T$ & \multicolumn{3}{c}{TGB} & \multicolumn{3}{c}{DRNet} & \multicolumn{3}{c}{LTRNet} & \multicolumn{3}{c}{LLTB}\\
% Dataset & $R_T$ & TGB &  DRNet & LTRNet & LLTB \\
\hline
Housing Price&  0.55 & \meduplo{36}{41}{28} & \meduplo{-}{}{} & \meduplo{62}{69}{55} & \meduplo{\textbf{22}}{24}{21}\\

Red Wine & 0.78 & \meduplo{43}{64}{32} & \meduplo{-}{}{} & \meduplo{116}{124}{110} & \meduplo{\textbf{9}}{12}{5}\\

Diabetes & 0.71 & \meduplo{41}{63}{23} & \meduplo{-}{}{} & \meduplo{113}{121}{101} & \meduplo{\textbf{7}}{9}{7} \\

Friedman & 0.46 & \meduplo{36}{37}{33} & \meduplo{-}{}{} & \meduplo{116}{133}{99} & \meduplo{\textbf{17}}{20}{16}\\

Car Price & 0.25 & \meduplo{25}{27}{22} & \meduplo{-}{}{} & \meduplo{33}{53}{26}  & \meduplo{\textbf{20}}{22}{18}\\
\hline
Breast Cancer & 0.19 & \meduplo{32}{41}{25} & \meduplo{$\infty$}{}{} & \meduplo{327}{373}{308} & \meduplo{\textbf{13}}{13}{7}\\

Banknote & 0.21 & \meduplo{30}{33}{26} & \meduplo{$\infty$}{}{} & \meduplo{32}{36}{30} & \meduplo{\textbf{6}}{6}{6}\\

Voice & 0.17 & \meduplo{17}{17}{17} & \meduplo{$\infty$}{}{} & \meduplo{168}{186}{156} & \meduplo{\textbf{6}}{9}{6}\\

Magic & 0.49 & \meduplo{43}{43}{33} & \meduplo{$\infty$}{}{} & \meduplo{110}{129}{98} & \meduplo{\textbf{19}}{23}{16}\\

Liver & 0.63 & \meduplo{37}{43}{32} & \meduplo{$\infty$}{}{} & \meduplo{81}{$\infty$}{28} & \meduplo{\textbf{18}}{27}{12}\\
\hline
\hline
Breast Cancer & 0.06 & \meduplo{$\infty$}{}{73} & \meduplo{20}{22}{17} & \meduplo{295}{346}{277} & \meduplo{\textbf{10}}{17}{7}\\

Banknote & 0.09 & \meduplo{28}{33}{24} & \meduplo{85}{104}{28} & \meduplo{22}{25}{18} & \meduplo{\textbf{6}}{5}{6}\\

Voice & 0.05 & \meduplo{52}{71}{29} & \meduplo{13}{15}{11} & \meduplo{104}{130}{90} & \meduplo{\textbf{4}}{4}{4}\\

Magic & 0.23 & \meduplo{42}{44}{32} & \meduplo{9}{10}{9} & \meduplo{66}{86}{60} & \meduplo{\textbf{6}}{7}{5}\\

Liver & 0.5 & \meduplo{8}{9}{7} & \meduplo{12}{16}{9} & \meduplo{21}{26}{14} & \meduplo{\textbf{7}}{9}{6}\\
\thickhline
\end{tabularx}
\end{table}

\providecommand{\inner}[2]{#1\trans#2} %{\langle #1, #2 \rangle} %{#1\trans #2}
\providecommand{\with}{\!\mid\,}
\providecommand{\zeronorm}[1]{\|#1\|_0}
\providecommand{\onenorm}[1]{\|#1\|_1}
\providecommand{\twonorm}[1]{\|#1\|_2}
\providecommand{\trans}{^\mathrm{T}}
\providecommand{\step}{\mathbf{1}}
\providecommand{\bq}{\mathbf{q}}
\providecommand{\bw}{\mathbf{w}}
\providecommand{\bteta}{\boldsymbol{\theta}}
\providecommand{\bx}{\mathbf{x}}
\providecommand{\bt}{\mathbf{t}}
\providecommand{\bg}{\mathbf{g}}
\providecommand{\by}{\mathbf{y}}
\providecommand{\bX}{\mathbf{X}}
\providecommand{\bbeta}{\boldsymbol{\beta}}
\providecommand{\bTeta}{\boldsymbol{\Theta}}
\providecommand{\bsigma}{\boldsymbol{\sigma}}
\providecommand{\bbR}{\mathbb{R}}
\providecommand{\setQ}{\mathcal{Q}}
\providecommand{\setW}{\mathcal{W}}
\providecommand{\setG}{\mathcal{G}}
\providecommand{\setI}{\mathcal{I}}
\providecommand{\loss}{\ell}
\providecommand{\Risk}{\mathcal{R}}
\providecommand{\obj}{\mathrm{obj}}
\providecommand{\bobj}{\mathbf{obj}}
\providecommand{\gb}{\mathrm{gb}}
\providecommand{\gs}{\mathrm{gs}}
\providecommand{\log}{\mathrm{log}}
\providecommand{\sqr}{\mathrm{sqr}}
\providecommand{\E}{\mathrm{E}}
\providecommand{\np}{k_{\mathrm{pr}}}
\providecommand{\ns}{k_{\mathrm{sp}}}
\providecommand{\halfspace}{\hspace{0.1em}}
\providecommand{\thickhline}{\noalign{\hrule height 0.25mm}}

\providecommand{\extratiny}[1]{\scalebox{0.8}{#1}}
\providecommand{\extratinytiny}[1]{\scalebox{0.72}{#1}}

\providecommand{\given}{\,|\,}
\providecommand{\meduplo}[3]{\extratinytiny{#3} & #1 & \extratinytiny{#2} }

\providecommand{\sign}{\mathrm{sign}}
\providecommand{\signg}{\zeta}
\newcolumntype{C}{>{\centering\arraybackslash}}
\newcolumntype{R}{>{\raggedright\arraybackslash}}
\newcolumntype{L}{>{\raggedleft\arraybackslash}}

\begin{table}[h!]
\centering
\caption{Test risk comparison of LLTBoost (LLTB), LTRNet, TGB, and DRNet methods on benchmark datasets using a complexity target (\text{$C_T$}) derived from TGB's median complexity presented in \Cref{table:complexity_table}. A squared loss for regression datasets, log loss and 0/1 loss for classification datasets have been computed as risk. Three numbers are reported for each experiment, representing the median (center), lower (left), and upper (right) bounds of the 93\% confidence interval.}\label{table:complexity_target_risk}
% \scriptsize
\begin{tabularx}{\textwidth}{Cm{0.15cm}CX|Cm{0.55cm}Lm{0.64cm}Cm{0.69cm}Rm{0.64cm}Lm{0.64cm}Cm{0.69cm}Rm{0.64cm}Lm{0.64cm}Cm{0.69cm}Rm{0.64cm}Lm{0.69cm}Cm{0.69cm}Rm{0.69cm}}
\thickhline
& Dataset & $C_T$ & \multicolumn{3}{c}{TGB} & \multicolumn{3}{c}{DRNet} & \multicolumn{3}{c}{LTRNet} & \multicolumn{3}{c}{LLTB}\\
\hline
\multirow{5}{*}{\rotatebox{90}{\scriptsize{Squared Loss}}} & Housing Price& 36 & \meduplo{0.54}{0.56}{0.53} & \meduplo{-}{}{} & \meduplo{$\infty$}{}{} & \meduplo{\textbf{0.49}}{0.51}{0.47}\\

& Red Wine & 43 & \meduplo{0.77}{0.81}{0.73} & \meduplo{-}{}{} & \meduplo{$\infty$}{}{} & \meduplo{\textbf{0.75}}{0.76}{0.73}\\

& Diabetes & 41 & \meduplo{0.73}{0.75}{0.65} & \meduplo{-}{}{} & \meduplo{$\infty$}{}{} & \meduplo{\textbf{0.62}}{0.66}{0.57} \\

& Friedman & 36 & \meduplo{0.45}{0.49}{0.42} & \meduplo{-}{}{} & \meduplo{$\infty$}{}{} & \meduplo{\textbf{0.36}}{0.37}{0.34}\\

& Car Price & 25 & \meduplo{0.23}{0.25}{0.21} & \meduplo{-}{}{} & \meduplo{0.28}{$\infty$}{0.25} & \meduplo{\textbf{0.22}}{0.24}{0.2}\\
\hline
\multirow{5}{*}{\rotatebox{90}{Log Loss}} & Breast Cancer & 32 & \meduplo{0.19}{0.2}{0.17} & \meduplo{0.31}{0.32}{0.31} & \meduplo{$\infty$}{}{} & \meduplo{\textbf{0.13}}{0.15}{0.1}\\

&Banknote & 30 & \meduplo{0.21}{0.23}{0.19} & \meduplo{0.42}{0.44}{0.41} & \meduplo{0.25}{$\infty$}{0.03} & \meduplo{\textbf{0.05}}{0.06}{0.05}\\

& Voice & 17 & \meduplo{0.16}{0.17}{0.15} & \meduplo{0.34}{0.35}{0.34} & \meduplo{$\infty$}{}{} & \meduplo{\textbf{0.13}}{0.14}{0.13}\\

& Magic & 41 & \meduplo{0.49}{0.5}{0.49} & \meduplo{0.56}{0.58}{0.55} & \meduplo{$\infty$}{}{} & \meduplo{\textbf{0.46}}{0.46}{0.45}\\

& Liver & 37 & \meduplo{\textbf{0.63}}{0.64}{0.63} & \meduplo{0.67}{0.7}{0.65} & \meduplo{0.64}{0.67}{0.62} & \meduplo{0.64}{0.66}{0.63}\\
\hline
\multirow{5}{*}{\rotatebox{90}{0/1 Loss}} & Breast Cancer & 32 & \meduplo{0.09}{0.1}{0.07} & \meduplo{0.05}{0.06}{0.05} & \meduplo{$\infty$}{}{} & \meduplo{\textbf{0.05}}{0.05}{0.03}\\

&Banknote & 30 & \meduplo{0.09}{0.1}{0.06} & \meduplo{0.1}{0.11}{0.09} & \meduplo{0.01}{$\infty$}{0.0} & \meduplo{\textbf{0.01}}{0.01}{0.01}\\

& Voice & 17 & \meduplo{0.07}{0.09}{0.05} & \meduplo{0.05}{0.05}{0.05} & \meduplo{$\infty$}{}{} & \meduplo{\textbf{0.04}}{0.04}{0.03}\\

& Magic & 41 & \meduplo{0.24}{0.25}{0.22} & \meduplo{0.22}{0.23}{0.21} & \meduplo{$\infty$}{}{} & \meduplo{\textbf{0.2}}{0.21}{0.19}\\

& Liver & 37 & \meduplo{\textbf{0.35}}{0.38}{0.34} & \meduplo{0.41}{0.43}{0.38} & \meduplo{0.40}{0.47}{0.32} & \meduplo{0.36}{0.38}{0.31}\\
\thickhline
\end{tabularx}
\end{table}

\providecommand{\inner}[2]{#1\trans#2} %{\langle #1, #2 \rangle} %{#1\trans #2}
\providecommand{\with}{\!\mid\,}
\providecommand{\zeronorm}[1]{\|#1\|_0}
\providecommand{\onenorm}[1]{\|#1\|_1}
\providecommand{\twonorm}[1]{\|#1\|_2}
\providecommand{\trans}{^\mathrm{T}}
\providecommand{\step}{\mathbf{1}}
\providecommand{\bq}{\mathbf{q}}
\providecommand{\bw}{\mathbf{w}}
\providecommand{\bteta}{\boldsymbol{\theta}}
\providecommand{\bx}{\mathbf{x}}
\providecommand{\bt}{\mathbf{t}}
\providecommand{\bg}{\mathbf{g}}
\providecommand{\by}{\mathbf{y}}
\providecommand{\bX}{\mathbf{X}}
\providecommand{\bbeta}{\boldsymbol{\beta}}
\providecommand{\bTeta}{\boldsymbol{\Theta}}
\providecommand{\bsigma}{\boldsymbol{\sigma}}
\providecommand{\bbR}{\mathbb{R}}
\providecommand{\setQ}{\mathcal{Q}}
\providecommand{\setW}{\mathcal{W}}
\providecommand{\setG}{\mathcal{G}}
\providecommand{\setI}{\mathcal{I}}
\providecommand{\loss}{\ell}
\providecommand{\Risk}{\mathcal{R}}
\providecommand{\obj}{\mathrm{obj}}
\providecommand{\bobj}{\mathbf{obj}}
\providecommand{\gb}{\mathrm{gb}}
\providecommand{\gs}{\mathrm{gs}}
\providecommand{\log}{\mathrm{log}}
\providecommand{\sqr}{\mathrm{sqr}}
\providecommand{\E}{\mathrm{E}}
\providecommand{\np}{k_{\mathrm{pr}}}
\providecommand{\ns}{k_{\mathrm{sp}}}
\providecommand{\halfspace}{\hspace{0.1em}}
\providecommand{\thickhline}{\noalign{\hrule height 0.25mm}}

\providecommand{\extratiny}[1]{\scalebox{0.8}{#1}}
\providecommand{\extratinytiny}[1]{\scalebox{0.72}{#1}}

\providecommand{\given}{\,|\,}
\providecommand{\meduplo}[3]{\extratinytiny{#3} & #1 & \extratinytiny{#2} }

\providecommand{\sign}{\mathrm{sign}}
\providecommand{\signg}{\zeta}
\newcolumntype{C}{>{\centering\arraybackslash}}

\begin{table}[h!]
\centering
\caption{Comparison of computational time (sec) for LLTBoost (LLTB), LTRNet, TGB, and DRNet methods on benchmark datasets.}\label{table:computational_time}
% \scriptsize
\begin{tabularx}{\textwidth}{CX|Cm{0.6cm}Cm{0.5cm}CXCXCXCX}
\thickhline
Dataset & d & n & TGB & DRNet & LTRNet & LLTB \\
\hline
Housing Price&  8 & 20640 & 0.88 & - & 10.65 & 1.69\\

Red Wine & 11 & 1599 &  1.15 & - & 11.38 & 1.47\\

Diabetes &10 & 442 & 0.86 & - & 10.77 & 1.53 \\

Friedman &10 & 2000 & 1.07 & - & 11.44 & 1.19\\

Car Price &4 & 1770 & 0.5 & - & 11.47 & 0.87\\
\hline
 Breast Cancer & 30 & 569 & 2.6 & 11.53 & 4.56 & 4.5\\

Banknote &4 & 1372 & 0.55 & 23.11 & 11.35 & 0.91\\

Voice &20 & 3168 &1.93 & 26.52 & 13.68 & 6.4\\

Magic &10 & 19020 & 1.1 & 34.26 & 14.08 & 2.43\\

Liver &6 & 345 & 0.47 & 19.03 & 10.47 & 0.93\\
\thickhline
\end{tabularx}
\end{table}

\section{Empirical Evaluation}\label{sec:evaluation}

We compare the risk/complexity trade-off of the proposed LLTBoost\footnote{\href{https://github.com/sherri1370/LinearTransRuleLearning}{https://github.com/sherri1370/LinearTransRuleLearning}} method with LTRNet, TGB and DRNet. In TGB, we use the standard gradient boosting algorithm introduced by \cite{dembczynski2010ender} and coded by \cite{yang2024orthogonal}\footnote{
\href{https://github.com/fyan102/FCOGB}{https://github.com/fyan102/FCOGB}}, optimizing the objective defined in \Cref{eq:obj_gb}. For DRNet, we used the code by authors at Github repository \footnote{
\href{https://github.com/Joeyonng/decision-rules-network/tree/master}{https://github.com/Joeyonng/decision-rules-network/tree/master}}.
For this comparison, we use ten common benchmark datasets, five for regression and five for classification as shown in \Cref{table:complexity_table,table:complexity_target_risk,table:computational_time}.
To ensure a consistent evaluation, we adopt the following evaluation protocol. All methods were run with ten repetitions, using ten random train/test splits based on bootstrap samples for the train and the corresponding out-of-bag samples for the test. Here, the size of the bootstrap samples was set to $n'=\min(n, 500)$, i.e., we worked with training size at most $500$ to avoid overly long computation times. 
To not underestimate the performance of the TGB baseline, the method was replicated with seven different regularization strengths $\{0.0001, 0.001, 0.01, 0.1, 1, 10, 100\}$ and for each dataset the best of these was considered (oracle performance). 
Similarly, to not underestimate the performance of the LTRNet baseline, the best result across the W-shaped penalties $\{0.0001, 0.001, 0.01, 0.1, 1\}$ were considered.
To realize different risk/complexity trade-offs, for TGB and the LLTBoost methods,
the maximum number of rules was varied over $r \in \{1, \dots, 10\}$. In contrast, for LTRNet and DRNet methods, this value was fixed at ten due to the specific assumptions underlying their approach.
Therefore, for these methods, a different parameter must be adjusted to control the complexity or the number of rules.
In LTRNet method, the strengths of the $L_1$ penalty terms were varied across $\{0.0001, 0.001, 0.01, 0.1, 1\}$ to control the number of rules and sparsity of linear transformations. Moreover, to ensure that the total computation time for one dataset remained within one hour, we set the number of gradient descent restarts to six and the maximum number of iterations to 250. For the final rounding step, we used 10\% of the maximum absolute value of each weight vector as the rounding threshold. 
According to DRNet achitecture, rules layer consists of ``AND'' gates representing the conjunctions of conditional terms. An $L_0$ regularization term with parameter $\lambda_1$ is applied to these gates, and we use six different $\lambda_1 \in \{0.0001, 0.001, 0.01, 0.1, 1, 10\}$. The other regularization term with parameter $\lambda_2$ for ``OR'' layer is fixed at $10^{-5}$ as specified in the original paper. All other parameters are set according to the original paper.
Additionally, for both LTRNet and LLTBoost methods, the maximum number of propositions, $k_r$ was set to 5.
In LLTBoost, the maximum number of non-zero weights for the sparse linear representation was set to 5.
All datasets were standardized as an essential pre-processing step for all methods.  

\subsection{Risk versus Complexity}
\Cref{table:complexity_table} compares complexities, as defined in \Cref{eq:c,eq:cq}, across benchmark datasets. The table reports the median minimum complexity to reach the test risk target ($R_T$) across 10 repetitions along with the 4th and 7th order statistics, which form a valid non-parametric 77\% confidence interval for the population median (see \cite[Ch.~5]{hahn2011statistical}).
If a repetition fails to meet the target, its complexity is recorded as ``$\infty$'', and methods with a median of this complexity are labeled as ``$\infty$''. In the top section (top ten rows), the test risk target is set based on the average test risk of TGB across its various complexity levels. As shown, by comparing the medians, LLTBoost achieves the lowest complexity across all datasets. DRNet is not applicable for regression datasets and failed to meet the risk target in classification datasets as well. Given that the primary focus of this method was on the comparison of 0/1 loss, we provide an additional complexity analysis in the second section of \Cref{table:complexity_table} (bottom five rows), using a 0/1 loss target derived from DRNet. This risk target is defined as the average test risk achieved by DRNet across all its complexity levels. As shown in the results, LLTBoost achieved the lowest complexities.
Additionally, \Cref{fig:risk_complexity_trend_banknote} illustrate the trends of risk with respect to complexity for the Banknote dataset. The black dashed line represents the target values defined in \Cref{table:complexity_table}.
Furthermore, \Cref{table:complexity_target_risk} presents a comparison of risk at a specified complexity target ($C_T$), determined based on the median complexity of TGB reported in \Cref{table:complexity_table}. For each repetition, the ensemble with the greatest complexity not exceeding this target was selected. In this table, a 93\% CI is used corresponding to 3rd and 8th order statistics. Consistent with \Cref{table:complexity_table}, the risk for repetitions that fail to meet $C_T$ is recorded as ``$\infty$'', and methods with a median risk at this level are labeled ``$\infty$''. Among the evaluated methods, LLTBoost consistently achieves the lowest test risk across all datasets and attains the lowest 0/1 loss in four out of five classification datasets.

\subsection{Computational Time}
\Cref{table:computational_time} presents the average computational time for one training run corresponding to one repetition and a single value for hyperparameters, such as penalty term strength. Notably, LTRNet's computational time depends on the number of restarts and iterations. A higher number of restarts and iterations increases the likelihood of finding a better optimum, but also results in higher computational costs.
 
\section{Conclusion}\label{sec13}

We introduced an extension to traditional rule ensembles by introducing learnable sparse linear transformations in the propositions, enabling decision regions with oblique faces and increasing model flexibility. Empirical results show that the proposed method significantly reduces the complexity of rule ensemble models compared to other baselines, while maintaining a favourable risk/complexity trade-off, albeit with potentially more complex data representations. These methods also reduce the need for manual feature engineering or expert domain knowledge. Future work could focus on introducing a more suitable search strategy to find the optimal number of propositions with a more sparse representation of input variables.

\begin{credits}
% \subsubsection{\ackname} A bold run-in heading in small font size at the end of the paper is
% used for general acknowledgments, for example: This study was funded

\subsubsection{\discintname}
The authors have no competing interests to declare that are relevant to the content of this article.
\end{credits}

\bibliographystyle{splncs04}
\bibliography{mybibliography}

\end{document}